\documentclass{article} %
\usepackage{iclr2027_conference,times}

\usepackage{amsmath,amsfonts,bm}

\def\eqref#1{equation~\ref{#1}}

\def\1{\bm{1}}

\DeclareMathAlphabet{\mathsfit}{\encodingdefault}{\sfdefault}{m}{sl}
\SetMathAlphabet{\mathsfit}{bold}{\encodingdefault}{\sfdefault}{bx}{n}

\usepackage{xcolor}
\definecolor{myred}{HTML}{D21418}
\usepackage{hyperref}
\hypersetup{colorlinks=true, citecolor=myred, linkcolor=myred, urlcolor=myred}
\usepackage{url}
\usepackage{amsmath}
\usepackage{graphicx}
\usepackage[nameinlink]{cleveref}
\usepackage{booktabs}
\usepackage{tcolorbox}

\newtcolorbox{thinkingquote}{
  colback=black!4,
  colframe=black!45,
  boxrule=0.4pt,
  arc=1pt,
  left=8pt,
  right=8pt,
  top=8pt,
  bottom=8pt,
}

\usepackage{xspace}
\newcommand{\methodname}{targeted guidance\xspace}
\newcommand{\Methodname}{Targeted guidance\xspace}

\title{Stabilizing On-Policy Targeted Guidance \\
in Agentic Reinforcement Learning}

\author{Kevin Frans\\
Cursor Research\\
UC Berkeley\\
\And
Alexander M. Rush \\
Cursor Research \\
\And
Sergey Levine\\
UC Berkeley\\
\And
Pieter Abbeel \\
UC Berkeley \\
\And
Ashvin Nair \\
Cursor Research \\
}

\newcommand{\redcolor}[1]{\textcolor{myred}{#1}}

\iclrfinalcopy %
\setcitestyle{numbers,square,comma} %

\begin{document}

\maketitle

\begin{abstract}
Naive on-policy RL is often bottlenecked by the capabilities of the base model --- insufficient coverage can result in an inability to learn from certain rewards. At the same time, rewards in modern post-training systems often come from rich textual sources, e.g. judge models with a rubric of criteria. Can we utilize such structures to provide a rich directional signal that enables improving on otherwise difficult to optimize rewards? In this work, we examine a framework we refer to as targeted guidance that learns from per-turn textual feedback via an on-policy self-distillation objective. Our focus is not only on directly improving hard-to-optimize behaviors, but on how to do so throughout the training process without degrading baseline RL performance. Over a range of settings across an agentic multi-turn software engineering environment, we find success by combining targeted self-distillation with a per-criteria RL reward, and by limiting feedback to turns where the policy specifically fails. We further examine the design choices that lead to stable entropy and reduce oscillatory behavior throughout training.
\end{abstract}

\section{Introduction}

The strength of large-scale reinforcement learning (RL) has been well established in recent years. However, there is often a limit to how much a policy can improve from naive on-policy RL~\citep{yue2025does}. The primary issue is that of \textit{insufficient coverage} -- even if a reward function is carefully selected to encourage desired behavior (e.g. use a new tool with the right format), the policy will nevertheless fail to improve if it does not have a chance of performing this behavior \citep{gandhi2025cognitive,zhao2025echo}. These problems are especially prominent in agentic settings, with multi-turn tool-calling, where a policy may be required to chain together \textit{many} complex actions in order to receive a reward signal. Fundamentally, while RL incentivizes existing good behaviors, such methods do not communicate \textit{how to perform new desirable behaviors}.

At the same time, we argue that sources of signal for communicating new behaviors are already present in many modern post-training systems. Rewards in such systems often come from a range of rich textual sources, e.g. judge models provided with a rubric of stylistic criteria. Naive RL methods condense the set of such signals into a single scalar reward, incurring a significant loss of information. Rather, can we utilize these existing sources to provide rich directional signal that allows us to learn from otherwise hard-to-optimize rewards?

In this work, we explore a framework we refer to as \textit{targeted guidance} which internalizes per-turn textual feedback via an expert distillation objective. The main idea is that each binary reward component can additionally define a textual feedback string. If a given rollout or agent turn fails to satisfy a given reward criteria, then alongside the typical policy gradient loss, we minimize KL towards an expert defined as a policy with the same underlying weights, but with the given feedback placed in context. For multi-turn settings, feedback is placed directly prior to the offending agent turn, and only that turn is affected by the distillation loss.

Our algorithmic setup builds on a rich body of prior work often referred to as on-policy self-distillation (OPSD) methods \citep{zhao2026opsd, shenfeld2026selfdistillation, hubotter2026reinforcement, song2026expanding}. In contrast to the one-shot benchmarks typically studied in such works, however, the main intent of this work is to study how such methods can be introduced directly to practical agentic RL systems with minimal regret to baseline RL performance. 
Naive on-policy distillation methods have previously been shown to cause phenomena such as reasoning shortening~\citep{kim2026selfdistillation,kaur2026rethinking,peng2026thinking} or diversity collapse~\citep{jin2026entropy,agarwal2024onpolicy}, both of which can lead to critical issues if affecting primary RL performance. We find that by combining a self-distillation objective with a per-criteria standard RL reward, and by limiting the distillation loss to the turns where the policy specifically fails, we are able to stabilize the learning process and broaden the policy capabilities \textit{while maintaining performance on primary RL tasks}.

\begin{figure}[t]
    \centering
    \includegraphics[width=\linewidth]{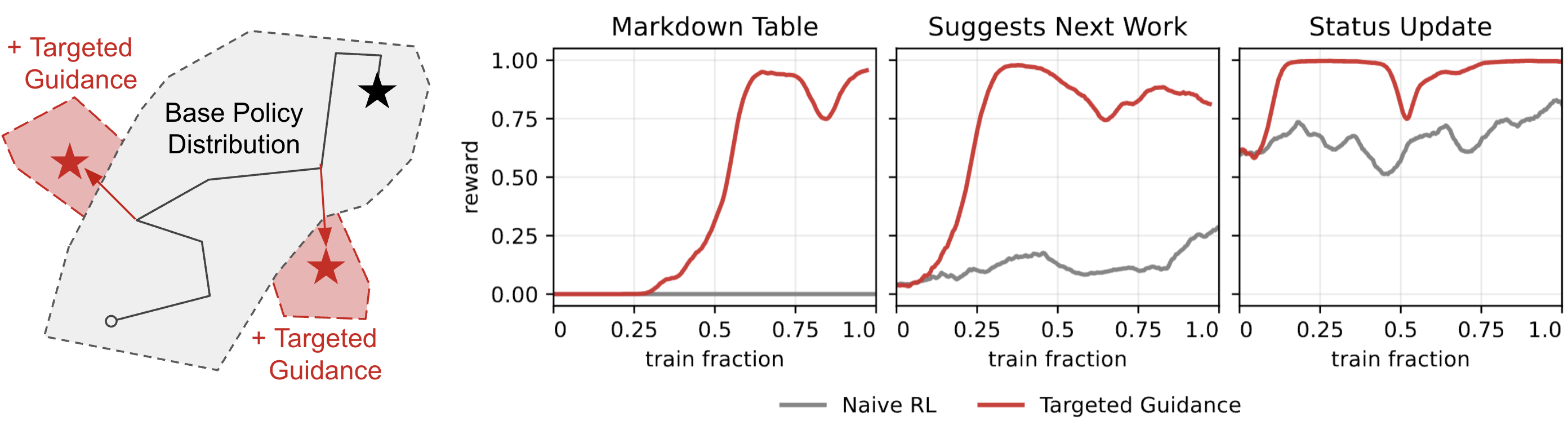}
    \caption{\textbf{Naive RL fails to optimize criterion that fall outside the support of base policies.} Exploration issues in RL can manifest as slow learning (middle) or in the worst case a complete failure to improve (left). On the other hand, \methodname enables policies to directionally improve specific aspects of their behavior and thus continue to learn.}
    \label{fig:nosignal}
\end{figure}

We empirically examine our methodology throughout an agentic setting in which a Qwen3-Coder-30B model\footnote{In this paper, all experiments use the publicly available Qwen3 model as a base. Cursor used a version of the techniques described here for the \href{https://cursor.com/blog/composer-2-5}{Composer 2.5 release}, which does not involve the Qwen series of models.} is post-trained to solve software engineering tasks through multi-turn harness involving tool calling and code editing. Notably our focus is not on improving coding pass-rate (which a well-tuned RL procedure can already do well), but rather on improving specific criterion in which naive RL procedures \textit{fail} to improve -- e.g. calling under-trained tool formats or writing with a specific style. We show that \methodname results in stable exploratory improvement even when multiple conflicting criterion are present, or a certain criterion must be satisfied across multiple turns. We find that by appropriately selecting the distillation weight and applying distillation to internal thinking traces, we are able to stabilize the entropy and reduce oscillations in the policy training trajectory.

\section{Related Work}

\textbf{Exploration in deep reinforcement learning.} 
Exploration has been a recurring theme of deep reinforcement learning, with a large body of work considering exploration from the perspectives of state-coverage~\citep{bellemare2016unifying,hazan2019provably,lee2019efficient,ecoffet2021first,liu2021behavior}, curiosity~\citep{pathak2017curiosity,burda2019exploration}, or skill discovery~\citep{gregor2016variational,eysenbach2019diversity,park2024metra,ajay2021opal,pertsch2020accelerating,frans2024unsupervised}.
A number of works have accordingly applied such methodologies to the language model RL setting~\citep{zhang2026count,song2025outcome,sun2025curiosity,dai2025cde,gao2025navigate,li2025jointly,qu2026rlad}.
More uniquely to language models, the idea of exploration via strong pretrained priors has been explored in forms such as self-reflection~\citep{shinn2023reflexion,madaan2023selfrefine,qu2024recursive,kumar2025training} or process rewards~\citep{uesato2022solving,lightman2024lets,wang2024mathshepherd,setlur2025rewarding}.
A solid body of work has explored how privileged information (e.g.\ human reference solutions) can be used as exploratory signal~\citep{yan2025luffy,nath2025adaptive,qu2026pope,fu2026ppm, chen2024language, zhou2026reinforcing, song2026expanding}.
Our work most similarly relates to the latter class of methods, in that we aim to utilize signal from privileged textual feedback to broaden the coverage of policies. Unlike most prior settings, we focus specifically on stabilizing such techniques when combined with agentic multi-turn RL.

\textbf{Distillation signal from expert policies.} 
It is well-known that strong experts can provide rich signal to improve downstream models ~\citep{hinton2015distilling,kim2016sequence}.
From an RL perspective, a body of works claim rich generalization can arise when distillation signal is be generated in an \emph{on-policy} manner ~\citep{ross2011reduction,agarwal2024onpolicy,gu2024minillm,lu2025onpolicy}.
In cases where a strong expert model is not available, prior works have examined the question of whether a strong policy constructed by re-using the same weights as the student, but with privileged information placed in the context ~\citep{askell2021general,snell2022learning,ye2026onpolicy,lopezpaz2016unifying,zhao2026opsd,shenfeld2026selfdistillation,penaloza2026privileged, hubotter2026reinforcement, song2026expanding}, often referred to as on-policy self-distillation (OPSD) methods. Our work builds off prior OPSD methodology and most closely relates to methods that encourage stability by gating distillation signal to failing rollouts \citep{li2026unifying, ding2026hdpo, lu2026self}. However, while these works consider using distillation to improve a single monolithic reward, our work instead aims to improve multiple auxiliary criterion (e.g. style, tool formats, etc) while preventing degradation in baseline RL performance.

\section{Background and Preliminaries}

\textbf{Markov decision process.} We define a Markov decision process (MDP) by a state space $S$, an action space $A$, an initial state distribution $p(s_0)$, a transition function $p(s'|s, a)$, and a reward $r(s)$. We define a parameterized policy $\pi_\theta(a|s)$ as a probability distribution of actions, which combined with the MDP creates a distribution of trajectories $p(\tau)$. We write the return on trajectories as $R(\tau)$.

\textbf{Language models as policies.} Throughout this work, we focus specifically on a policy class defined by a neural network language model. We assume access to a base model which defines a starting policy $\pi_0$. For the agentic use case, the base model is generally a pretrained language model that has gone through a supervised finetuning (SFT) phase to adapt it to the tool call setting. As this work explicitly considers multi-turn environments with agents, we abstract away from individual tokens and refer to actions as an entire set of autoregressively generated tokens in sequence, e.g. a specific tool call. We assume that the Markovian state is fully defined by the sequence of previous observations and actions and their underlying tokens.

\textbf{Rewards and judges.} The reward function for this system $R(\tau)$ is nonzero only at the end of the rollout. 
We assume that the reward is a weighted additive function $R(\tau) = \sum_i \lambda_i R_i(\tau)$ of a prespecified set of criteria. 
The primary component of the reward is verification of task completion, but there are a collection of other properties such adherence to stylistic guides or use of specific tools. We will assume these rewards properties are mostly independent, e.g. it is possible to modify style without impacting task completion. While not central to the method, we primarily focus on the case of judge-based rewards, where each criterion $R_i$ is implemented as a call to judge LLM. This judge is equipped with additional privileged information on the correct way to solve the problem, and in addition to producing reward scores, can provide textual feedback justifying its result.

\section{Textual Feedback as a Targeted Directional Signal}

\begin{figure}[t]
    \centering
    \includegraphics[width=\linewidth]{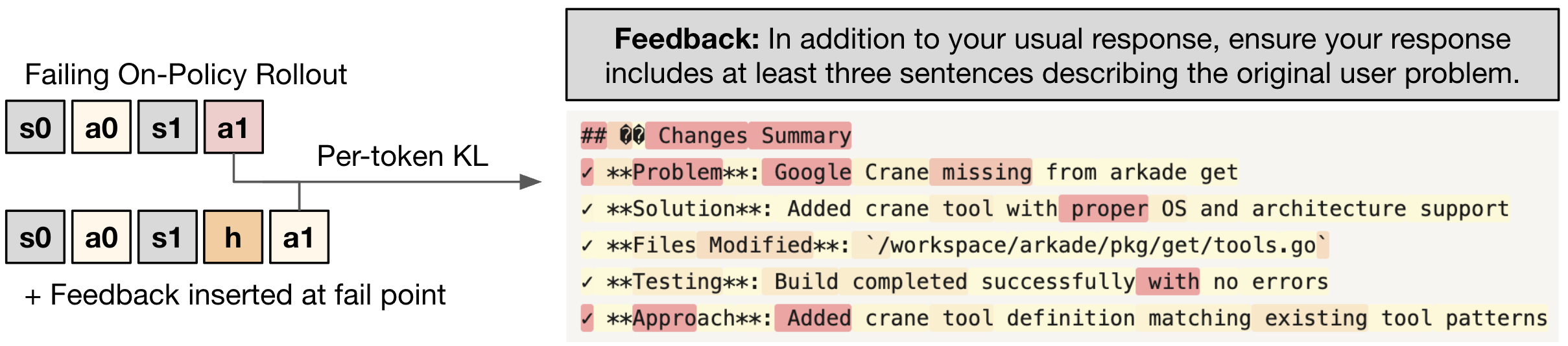}
    \caption{\textbf{\Methodname derives per-token signal from per-turn hints.} During the course of a standard RL procedure, in cases where rollouts fail to satisfy certain criteria, e.g. to utilize a certain writing style, textual feedback is employed to guide the policy towards the solution. We instantiate this via a per-token KL loss against a feedback-conditioned expert policy. }
    \label{fig:method}
\end{figure}

As illustrated in \Cref{fig:nosignal}, it is often the case that naive policy gradients will fail to optimize even simple criterion, if the actions that satisfy such criterion are not within the support of the base policy. In the following section, we describe a strategy to construct a source of \textit{directional} policy improvement signal, which communicates not which rollouts were successful, but rather \textit{how to modify} specific aspects of a policy's behavior. 

Our main approach is to utilize a per-token distillation target to provide this directional signal (\Cref{fig:method}). Rather than returning a single scalar, each reward component can return \textit{textual feedback} in the form of a feedback string $f$ associated with a specific turn of the agent rollout. 
Inserting $f$ before that turn defines an expert that shares the current policy's weights, $\mu(\cdot\mid s) = \pi_\theta(\cdot\mid s, f)$ at targeted turns and $\mu = \pi_\theta$ elsewhere. Following what is often referred to as on-policy self distillation \citep{zhao2026opsd}, we minimize KL between the live policy and $\mu$ over the on-policy state distribution $d^\pi$, treating $\mu$ as fixed within an update.

\textbf{Expert distillation as approximate policy improvement.} We base our algorithmic considerations on classical policy improvement arguments \citep{kakade2002approximately, schulman2015trust, achiam2017constrained}. A policy $\mu$ provides a useful improvement direction if its expected advantage on the current policy's state distribution is positive:
\begin{equation}
    \mathbb{E}_{s\sim d^\pi}\,\mathbb{E}_{a\sim\mu(\cdot\mid s)}\big[A^\pi(s,a)\big] > 0,
    \label{eq:advantage}
\end{equation}
in which case a sufficiently small step towards $\mu$ results in an improved policy. In the case where distillation is understood to specify how certain behaviors (e.g. code style) can improve, but should not affect other behaviors (e.g. code behavior), we in general prefer experts that limit their differences from the base policy, i.e.
\begin{equation}
    \mathbb{E}_{s\sim d^\pi}\Big[\mathrm{KL}\big(\pi(\cdot\mid s)\,\|\,\mu(\cdot\mid s)\big)\Big] < \delta .
    \label{eq:kl}
\end{equation}
While neither condition is guaranteed in practice, we use
these as design heuristics: feedback should result in an expert better than the base policy, and the expert should otherwise remain as close to the policy as possible.

\textbf{Stabilizing the targeted guidance update.} The aim of our proposed framework is to integrate targeted distillation-based guidance into agentic RL in a way that balances exploratory pressure across targeted facets while ensuring that baseline RL performance is not affected. Combining the approaches naively leads to some risk of policy collapse or stagnation. We therefore introduce two simple ideas that stabilize the training procedure, as described below.

The first main idea is that we would like to minimize the amount of unnecessary pressure from the distillation objective, which can be heuristically understood as tightening the divergence between the expert and live policy (\Cref{eq:kl}). Thus, in contrast to prior work which places privileged information into a global context to construct the self-distillation target, we place the feedback string right before the affected agent turn, and mask out the KL loss on all other turns. Additionally, the distillation objective is \textit{only evaluated on agent turns that contain a failed criterion.} 

The second core idea is that while self-distillation can provide a strong directional signal of improvement, it is less clear that such objectives result in stable convergence. For this reason, our proposed methodology is to combine the distillation signal with a standard RL reward on each desired criteria. The intuition is that distillation serves as a targeted way to ensure the policy learns the relevant behavior, while it is the unbiased RL objective that enables the policy to retain and refine such behaviors.

The distillation loss in our setting is calculated via reverse KL distance between the student and the teacher. In our experiments, we found that forward-KL, Jensen-Shannon distance, and Rao-Blackwellized KL \citep{amini2026better} performed similarly, although an in-depth study was not performed. To reduce undesirable effects from inference-trainer mismatch, we utilize a full-vocabulary KL loss computed entirely on the trainer side, rather than a sample-based estimator. The combined objective is therefore:
\begin{equation}
  \mathcal{L}(\theta) = \mathcal{L}_{\text{GRPO}}(\theta)
  + \lambda \sum_{t \in \mathcal{F}}
  \mathrm{KL}\big(\pi_\theta(\cdot \mid s_t) \,\|\, \text{sg}(\pi_\theta(\cdot \mid s_t, f_t))\big),
  \label{eq:combined}
\end{equation}
where $\mathcal{F}$ is the set of agent turns that fail a criterion, $f_t$ is the
feedback for turn $t$, $\lambda$ is the distillation weight. The `sg` operator represents stop gradient. Additionally, we normalize the distillation KL term across all tokens present in a given rollout, reducing the impact of long rollouts with high KL distances, which often correspond to undesirable states such as looped reasoning.

\textbf{How should one design the textual feedback content?} Recall that an important criterion for expert guidance to be an effective improvement operator is that the expert itself must have higher expected reward (\Cref{eq:advantage}). As experts in our case are defined as the same model but with certain information in the context, we can iterate on this context via a simple procedure -- query the base model with a considered feedback string, measure average performance against the contextless model, and only allow prompts which show an average improvement. Note that although we evaluate the feedback via autoregressive generation in this procedure, the actual application of this feedback is for per-token on-policy distillation.

\textbf{Is \methodname asymptotically unbiased?} While auxiliary signals can be necessary to solve hard exploration problems, they can at times prevent policies from converging to the optimal policy. As an example, it is possible that $\mathrm{KL}\big(\pi_\theta(\cdot \mid s_t) \,\|\, \pi_\theta(\cdot \mid s_t, f_t)\big)$ is nonzero even for an optimal policy due to spurious affects of conditioning on feedback. For this reason, we instead argue that by gating the distillation loss to only affect failing turns, the loss function simplifies to a simple policy gradient loss in the case that all feedback criterion are consistently satisfied, and our procedure will converge to the optimal policy assuming that such a policy exists.

\textbf{What is the computational cost of \methodname?} For many practical RL systems, the main computational cost is in sampling rollouts rather than training on them. A benefit of the on-policy distillation formulation is that the teacher is used purely for logit-level guidance and does not require autoregressive generation. The memory requirements of full-logit distillation can be alleviated via kernels that materialize only the last hidden layer representation \citep{hsu2025ligerkernel, quack2025}. In this way, the main cost of adding targeted guidance to a standard RL procedure is a single additional forward pass to gather expert logit targets.

\section{Empirical Experiments with \Methodname }

\subsection{Didactic Example: Tool Format Correction}

\begin{figure}[t]
    \centering
    \includegraphics[width=\linewidth]{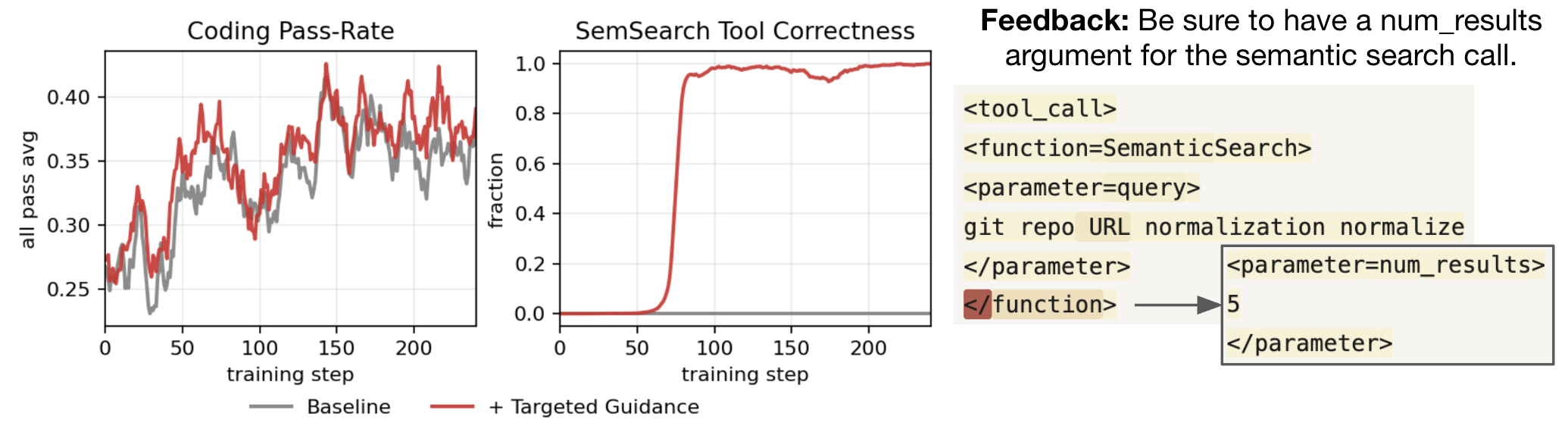}
    \caption{\textbf{Didactic example: \Methodname teaches new tool call formats.} In cases where the base policy fails to fulfill certain criterion, e.g. to utilize a certain tool call format, textual feedback is utilized to guide the policy towards the solution. We instantiate this via a per-token KL loss against a feedback-conditioned expert policy. As shown on the right \redcolor{in red}, only the token in which the student and expert diverge (in this case, the tool format delimiter) receives a strong signal. 
    }
    \label{fig:didactic}
\end{figure}

As a motivating example, we examine an auxiliary criterion that agent policies struggle with: learning to use new tools. Oftentimes, one may want to introduce new tools or tool arguments during the RL phase, even when the base model has already been through a supervised fine-tuning phase that did not include this tool. We show now that \methodname provides a natural way to address this problem without performance regression.

Our specific setup is the optimization of a multi-turn agentic coding policy, in which an agent utilizes a series of tool calls to solve a given problem, as evaluated by hidden test cases. We train such a model with standard policy gradient methods, assigning each rollout a weighted sum of pass-rate, tool format, and length-penalty rewards. We consider the setting in which we are introducing a new tool-calling format to the ``semantic search" tool. This format involves specifying an additional parameter: the number of lines that the search should return. Naively, the base model fails to utilize this new format at all, even when example usage is placed in the prompt, in part due to SFT not including this new format.

To fix this behavior, we can utilize \methodname. During training, whenever the agent uses the semantic search tool without the correct format, we inject a feedback string right before the turn with the error. The policy conditioned on the feedback string then becomes an expert to serve as a source of guidance signal, which we combine with the standard policy gradient loss during RL training.

As shown in \Cref{fig:didactic}, a policy trained with \methodname learns the new tool calling format, without suffering a regression on the primary coding performance. We attribute this to the guidance signal being targeted and minimal -- as shown on the right, only tokens which show strong disagreement to the expert receive a large signal. Importantly, guidance is always done on on-policy rollouts exactly where the error occurs, to reduce any spurious distillation effects. As the model fixes the undesired behavior, the guidance signal accordingly drops to near zero.

\subsection{Can \Methodname work when considering a suite of criteria?}

\begin{table*}[t]
\centering
\caption{A comparison of methods that aim to learn a suite of stylistic criteria (main columns), while maintaining primary coding pass-rate (right column). Errors represent standard deviation over the last 32 steps of training.}
\vspace{10pt}
\label{tab:method_comparison}
\setlength{\tabcolsep}{4pt}
\setlength{\aboverulesep}{0pt}
\setlength{\belowrulesep}{0pt}
\renewcommand{\arraystretch}{1.2}
\resizebox{\textwidth}{!}{%
\begin{tabular}{l|ccccccc|c}
\toprule
Method & Markdown & Sol. Desc & Req. Order & Approval & Bolded & Cmd Ran & Suggest Work & Pass Rate \\
\midrule
Baseline   & $0.00_{\pm .00}$ & $0.70_{\pm .06}$ & $0.04_{\pm .03}$ & $0.00_{\pm .00}$ & $0.43_{\pm .07}$ & $0.00_{\pm .00}$ & $0.03_{\pm .03}$ & $\mathbf{0.33}_{\pm .05}$ \\
GRPO       & $0.00_{\pm .00}$ & $\mathbf{0.96}_{\pm .02}$ & $0.48_{\pm .08}$ & $0.00_{\pm .00}$ & $\mathbf{0.95}_{\pm .01}$ & $0.00_{\pm .00}$ & $0.20_{\pm .07}$ & $\mathbf{0.33}_{\pm .07}$ \\
OPSD       & $0.77_{\pm .04}$ & $0.88_{\pm .03}$ & $0.29_{\pm .03}$ & $0.25_{\pm .09}$ & $0.90_{\pm .02}$ & $0.01_{\pm .00}$ & $0.44_{\pm .04}$ & $0.30_{\pm .05}$ \\
Online SFT & $0.01_{\pm .01}$ & $0.78_{\pm .03}$ & $0.29_{\pm .03}$ & $0.02_{\pm .01}$ & $0.69_{\pm .03}$ & $0.00_{\pm .00}$ & $0.28_{\pm .07}$ & $0.26_{\pm .05}$ \\
LaTRO      & $0.00_{\pm .00}$ & $0.57_{\pm .19}$ & $0.03_{\pm .02}$ & $0.00_{\pm .00}$ & $0.21_{\pm .11}$ & $0.00_{\pm .00}$ & $0.07_{\pm .04}$ & $0.14_{\pm .09}$ \\
Verifree   & $0.01_{\pm .02}$ & $0.83_{\pm .04}$ & $0.26_{\pm .14}$ & $0.01_{\pm .01}$ & $0.65_{\pm .19}$ & $0.00_{\pm .00}$ & $0.15_{\pm .11}$ & $0.20_{\pm .06}$ \\
\textbf{TG (ours)} & $\mathbf{0.85}_{\pm .09}$ & $\mathbf{0.95}_{\pm .01}$ & $\mathbf{0.77}_{\pm .05}$ & $\mathbf{0.55}_{\pm .06}$ & $\mathbf{0.99}_{\pm .00}$ & $\mathbf{0.64}_{\pm .13}$ & $\mathbf{0.85}_{\pm .03}$ & $\mathbf{0.33}_{\pm .05}$ \\
\bottomrule
\end{tabular}%
}
\end{table*}

We now examine applying \methodname in a more complex setting, in which we desire to optimize a suite of criteria, some of which may conflict with each other. Specifically, we focus on the problem of improving the communication style of coding agents. We first define a set of binary stylistic criterion to be verified by an external judge, such as ``include a markdown table" or ``describe the user problem". For each criterion, we define a binary reward function along with an associated static feedback string.

The aim is to enable policies to reliably satisfy the stylistic criterion, while also maintaining performance levels on the primary coding tasks. In particular, our goal is that policies trained via targeted guidance should explore, while also not showing any degradation on coding pass-rate as compared to a baseline policy trained without any style criterion at all.

Our base policy is a Qwen3-Coder-30B model further SFT'd for agentic coding in our harness. In all methods below, we combine various style-optimization methodologies with a baseline GRPO algorithm that optimizes for coding pass-rate and regularizes total output length. We train for 200 steps with a batch size of 32 groups with 32 rollouts each, using Adam and a learning rate of 3e-6. The training dataset is a single-epoch set of software engineering problems, i.e. the same prompt is only ever trained on once. We utilize an asynchronous RL pipeline~\citep{piche2025pipelinerl} where the inference policy may be older than the currently trained policy. For all distillation-style methods, we implement the same distillation operator of a full KL distance over the entire vocabulary, and we include both thinking and tool call tokens in the loss. We sweep the distillation weight and RL style reward weight hyperparameters.

We compare against a variety of representative methods:
\begin{itemize}
    \item \textbf{Baseline} represents a policy that only optimizes the standard coding reward. This serves as a baseline for the expected coding pass-rate when we do not take style criteria into account.
    \item \textbf{GRPO} \citep{shao2024deepseekmath} shows the performance of a standard RL policy where each stylistic criteria is assigned a binary reward, whose sum is optimized as part of the standard GRPO objective.
    \item \textbf{OPSD} \citep{zhao2026opsd}, a representative on-policy distillation method, which makes use of the per-criterion feedback to provide distillation signal. Unlike our proposed method, no additional RL reward is used beyond the baseline.
    \item \textbf{Online SFT}, a representative iterated SFT method, in which the feedback is used to construct an explicit teacher policy which autoregressively generates alternative teacher turns to replace turns that fail a criterion. These alternative turns are then used as SFT targets for the live policy during the RL process.
    \item \textbf{LaTRO} \citep{chen2024language}, a representative likelihood-as-reward method. Similarly to online SFT, we construct alternative turns from a teacher. The log-likelihood of the alternative turns are then assigned as a reward for all previous turns in the rollout.
    \item \textbf{Verifree} \citep{zhou2026reinforcing}, which we instantiate as using the same framework as LaTRO except using the raw likelihood as the reward.
    \item \textbf{Targeted Guidance (ours)}, in which both GRPO and distillation are combined. A distillation loss is only applied to turns in which a nonzero set of criterion are unfulfilled.
\end{itemize}

As shown in \Cref{tab:method_comparison}, policies trained with \methodname\ are able to improve on a wide range of evaluation criteria, while maintaining a coding pass-rate similar to that of standard GRPO. Naive OPSD maintains a relatively healthy coding pass-rate, however, the improvements on style criteria are less consistent, resulting in a worse final performance.

On the other hand, the online SFT, Verifree, and LaTRO policies result in a non-trivial degradation in coding pass-rate. In the case of online SFT, this may be because the continual SFT loss during RL limits how much the policy can change at each step, especially in an asynchronous setting. Verifree and LaTRO result in a further degradation in coding pass-rate. We hypothesize that the instability of these methods comes from the fact that unlike the on-policy methods, these three rely on \textit{sampling entirely new actions from the feedback-conditioned expert}. While increasing the likelihood of such actions may help in terms of style, such actions may have unintended consequences in terms of primary coding pass-rate. Thus, it is likely that instability arises from increasing likelihood of potentially suboptimal actions, especially in an asynchronous regime where such actions may become stale.

\subsection{Does \Methodname remain stable even when optimizing multi-turn interleaved feedback?}

\begin{figure}[t]
  \includegraphics[width=\linewidth]{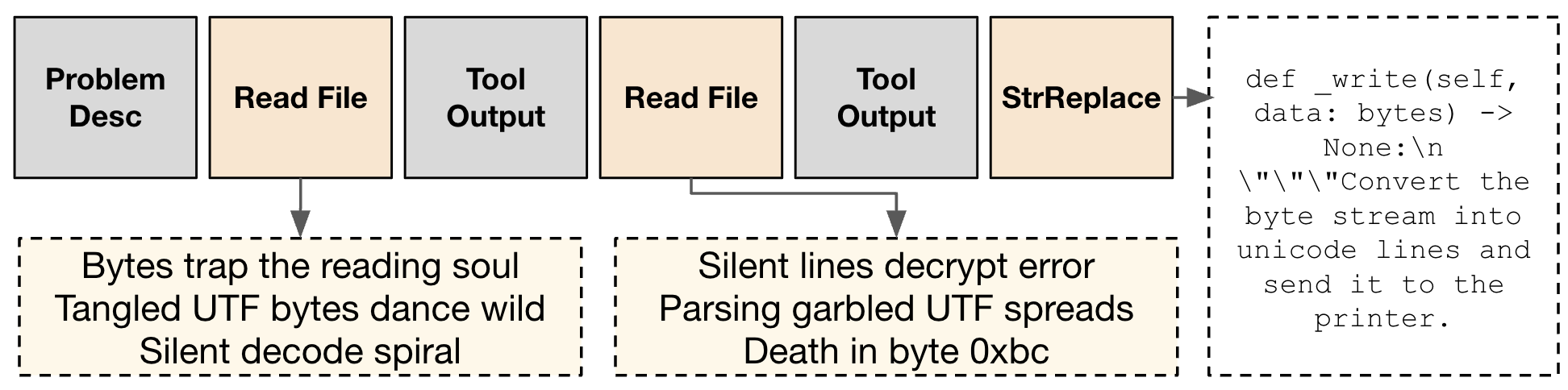}
  \caption{\textbf{A haiku-style poem produced by a policy trained via targeted guidance.} Feedback is inserted at every turn in which the desired criterion is not achieved.}
  \label{fig:encoding-haiku}
\end{figure}

\begin{figure}[t]
    \centering
    \includegraphics[width=\linewidth]{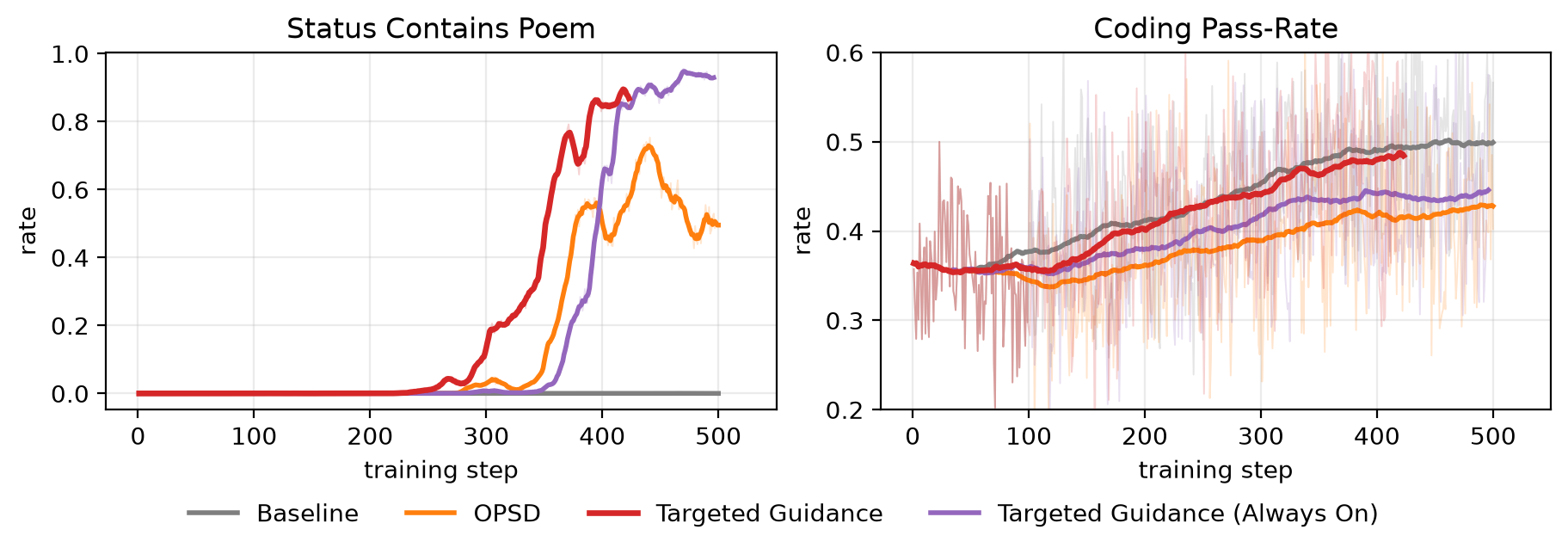}
    \caption{\textbf{Targeted guidance and alternatives as used to train models that for each agent turn report status updates as a poem.} The baseline GRPO run fails to optimize this out-of-support objective entirely. Naive OPSD learns, but is unstable. A variant of targeted feedback where feedback is \textit{always} provided (rather than only on failing turns) results in a regression in coding pass-rate.}
    \label{fig:poem}
\end{figure}

The above experimental setting considered stylistic criteria on the final message style, which only occurs at the end of an agentic rollout. We are also interested in improving the behavior of non-final actions that occur throughout an agent's rollout. 

To examine this, we construct a setting where \textit{each} agent turn is graded on verifiable auxiliary criteria. To study this setting, we adopt a criterion that requires significant deviation from the base model behavior: each turn must include a status report in the form of a poem. This is an out-of-distribution behavior that lies outside the support of the base model.

\Cref{fig:poem} showcases the learning trajectories of \methodname and baseline methods under the every-turn criterion setting. The GRPO baseline here results in a consistent coding pass-rate improvement, as expected, but completely fails to optimize the poem reward criterion. In comparison, the policy trained with \methodname is able to adopt the poem output style while minimizing regression on coding performance.

Ablations on two aspects of the framework reveal their necessity. In the OPSD run, where expert distillation guidance is applied but not combined with a corresponding reward for poem-correction, the model does learn to output poems at a reasonable rate, but the policy itself is unstable and fails to consistently hold this behavior. In addition, there is a significant regression to coding pass-rate. A similar regression also occurs when utilizing a variant in which the poem feedback is \textit{always} applied, regardless of whether that particular rollout was successful or not, highlighting the necessity of gating distillation only to failing turns.

\begin{figure}[t]
    \centering
    \includegraphics[width=\linewidth]{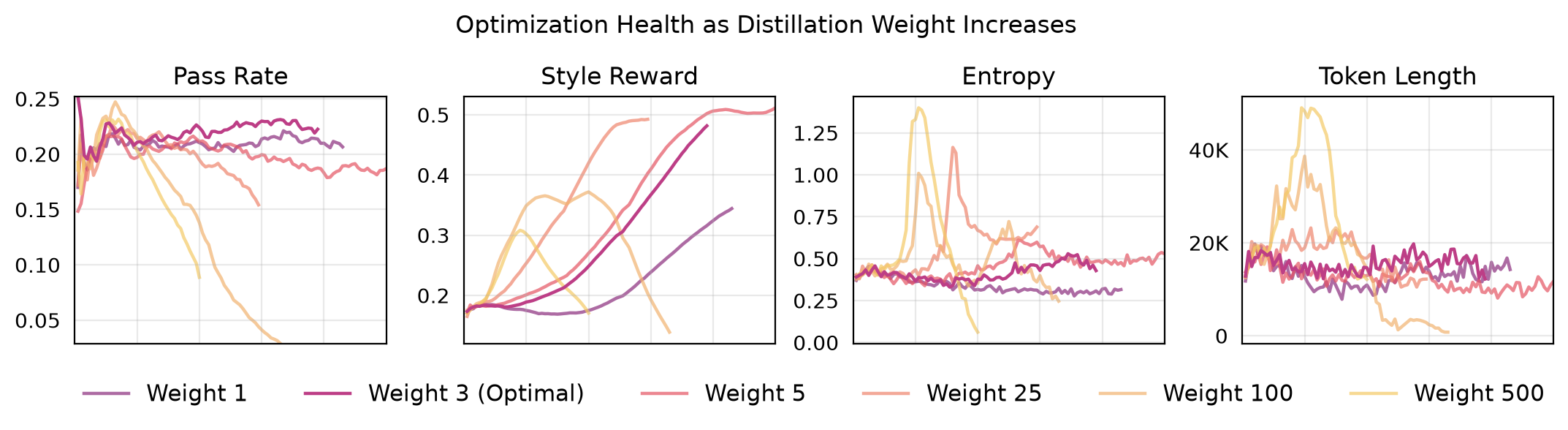}
    \caption{\textbf{Entropy spikes can occur if distillation weight is too high.} In our setting, this results in a spike in average token length, followed shortly by a collapse in pass-rate. With low enough distillation weight, the policy is stable throughout RL and style rewards can continue to improve without coding pass-rate degradation.}
    \label{fig:entropy}
\end{figure}

\begin{figure}[t]
    \centering
    \includegraphics[width=\linewidth]{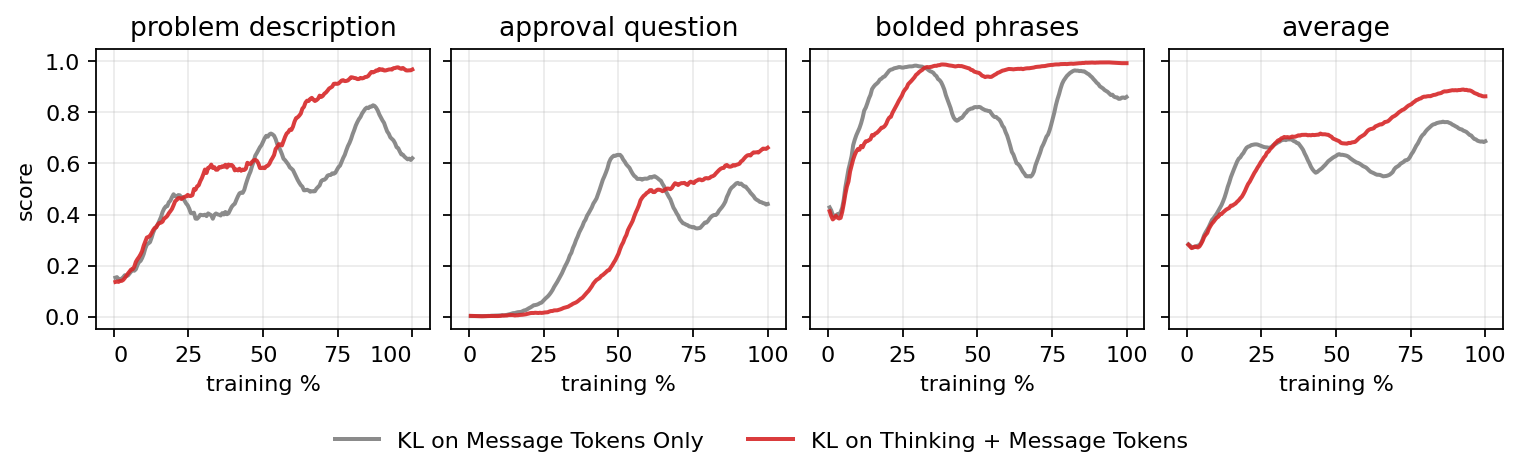}
    \caption{\textbf{Including thinking tokens in the distillation objective is crucial for stability.} In cases where the thinking is excluded from distillation signal, style criterion tend to display oscillatory behavior where certain criterion may improve, but later fall back down.}
    \label{fig:thinking}
\end{figure}

\subsection{What are the failure cases of the method, and how can they be alleviated?}

\textbf{Stabilizing entropy.} We find that an early predictor of collapse when utilizing distillation signal is a sharp increase in policy entropy. In our setting, this is accompanied with a steady increase in the average output length, and is quickly followed by a decrease in the coding pass-rate. 

An increase in entropy from exploratory signal is expected, to some degree. As by nature the behaviors that require exploratory signal are low likelihood under the live policy, increasing the likelihood of such behaviors will cause an increase in entropy. The issue is not entropy increase in general, but rather a spike in entropy that implies an over-correction or collapse. One possible explanation of the severity of such spikes is due to the recursive nature of self-distillation -- not only does the student increase in entropy, but the expert distribution may also increase as both policies share the same underlying weights. Continual self-distillation in this manner can thus result in a sharp spike as entropy increase is consistently reinforced.

As shown in \Cref{fig:entropy}, we can replicate entropy spiking settings by increasing the weight of the distillation signal. While high distillation weight runs display a higher style reward at the start, they quickly collapse and end up failing in both pass rate and in style criterion. On the other hand, lower distillation weights do not show this type of instability, but in turn improve on style at a slower rate. We find that a static parameter is enough to remain in the stable regime, but further efficiency may be gained via dynamic weighting \citep{levine2013guided, shenfeld2023tgrl}.

\textbf{Stabilizing intra-reward oscillations.} A phenomenon that appears when examining the set of stylistic rewards is that the policy oscillates between satisfying a subset of the criterion, improving on some criterion while degrading on others. We hypothesize this behavior arises from the competing pressure for all stylistic rewards -- when a distillation signal is present to push the model towards a certain style, the model may in turn be disincentivized to use a different style.

We find that such oscillatory behavior becomes much less noticeable when the distillation signal is applied also to the thinking tokens before the final agent message. Without distillation pressure on the thinking tokens, the policy shows highly oscillatory behavior. When thinking token KL is enabled, the behavior is instead a stable curve and increasing monotonically (\Cref{fig:thinking}).

To examine this phenomenon qualitatively, we look at the thinking traces for a model trained with and without distillation pressure on the thinking tokens. In the absence of distillation on thinking, the thinking trace for the final message does not mention style at all:

\begin{thinkingquote}
Okay, I've modified the code to support arrays of state patterns. The original issue was that the user was trying to use multiple conditions with the same key ('active'), which doesn't work in JavaScript objects because duplicate keys aren't allowed.
\end{thinkingquote}

On the other hand, if distillation is applied to the thinking tokens as well, then the thinking content of the final message includes a range of self-reminders to include specific stylistic components:

\begin{thinkingquote}
Now I should \redcolor{provide a clear summary} to the user about what I did. I should \redcolor{explain the issue, the solution}, and \redcolor{ask for their confirmation} to proceed or just implement it.
I should \redcolor{format this nicely with proper markdown}.
\end{thinkingquote}

As shown in \Cref{fig:thinking}, by distilling feedback into the thinking tokens, policies can incorporate a variety of feedback sources in a compositional manner, reducing oscillation and instead learning to follow the entire set of criterion.

\section{Discussion and Conclusion}

In this work, we introduced a simple \methodname approach that combines textual feedback self-distillation with standard on-policy RL.  
This allows for optimization of rewards criteria that may be difficult or impossible to optimize with policy gradients alone, while remaining stable enough to not interfere with baseline RL performance.

\textbf{Limitations.} One limitation of our proposed framework is that each reward in the suite requires a designed static feedback string. Follow-up work may examine whether strong feedback can be automatically learned, perhaps via a tied RL process. Such processes may also see whether a gating process or weighting between RL signal and distillation signal can be automatically learned.

Experimentally, the settings studied in this work focus solely on the multi-turn software engineering setup, and our targets for guidance largely revolve around style and formatting rather than primary intelligence. We note that there exist prior works which span a wide set of OPSD variants 
\citep{zhao2026opsd,shenfeld2026selfdistillation,penaloza2026privileged, hubotter2026reinforcement, song2026expanding, li2026unifying, ding2026hdpo, lu2026self}
which we were unable to port and compare to head-to-head under our experimental setting. A breakdown of which atomic design choices matter in the style of \citep{huang202237, engstrom2020implementation, andrychowicz2020matters} would likely be of interest.

\textbf{Future work.} Central to this method is the use of language as 
communicative medium between the reward and the policy, i.e. the reward judge needs a method for producing targeted textual feedback 
and the teacher model needs to condition on this information in an effective manner. In this work, we rely on the zero-shot ability of the policy to process this information and human prompting for the judge to produce effective feedback. We anticipate that further SFT on the policy can improve the effectiveness of distillation signal, and that the reward judges themselves could be fine tuned to produce more targeted hints that minimize reward interference. Both operations above could potentially be optimized via an interleaved meta optimization loop.

The approach in this work focused specifically on a per-token distillation approach to utilizing feedback. However, there may be strong benefits from allowing the teacher to autoregressively generate alternative turns, or otherwise rewrite the on-policy rollouts.
Recent work on teacher generation approaches, such as pedagogical RL~\citep{chakraborty2026pedagogicalrl}, suggest that training teacher models to provide guidance that is closer to the student distribution might be an effective way forward.

Finally, the feedback sources in this work focuses on targeted guidance at the turn-level. This approach fits with the reward criteria that were targeted for this work. A natural question is whether a similar approach could be used for dense guidance on the full rollout. While RL is a strong baseline in this setting, these results indicate that with the right stabilization text feedback may be able to complement with global exploration. 

\section*{Acknowledgements}

We thank the Cursor research team for their helpful feedback and input on this work. This work was supported in part by an NSF Fellowship for KF, under grant No.\ DGE 2146752, and AFOSR FA9550-22-1-0273 and ONR N00014-22-1-2773 for SL. Any opinions, findings, and conclusions or recommendations expressed in this material are those of the author(s) and do not necessarily reflect the views of the NSF. PA holds concurrent appointments as a Professor at UC Berkeley and as an Amazon Scholar. This paper describes work performed at Cursor and UC Berkeley, and is not associated with Amazon.

\bibliography{iclr2026_conference}
\bibliographystyle{iclr2027_conference}

\end{document}